\documentclass[conference]{IEEEtran}
\IEEEoverridecommandlockouts
\usepackage{cite}
\usepackage{amsmath,amssymb,amsfonts}
\usepackage{algorithmic}
\usepackage{graphicx}
\usepackage{textcomp}
\usepackage{xcolor}
\usepackage{bbding}
\def\BibTeX{{\rm B\kern-.05em{\sc i\kern-.025em b}\kern-.08em
    T\kern-.1667em\lower.7ex\hbox{E}\kern-.125emX}}

\makeatletter
\newcommand{\linebreakand}{%
  \end{@IEEEauthorhalign}
  \hfill\mbox{}\par
  \mbox{}\hfill\begin{@IEEEauthorhalign}
}
\makeatother

\begin{document}

\title{Mutual Query Network for Multi-Modal Product Image Segmentation
}

\author{\IEEEauthorblockN{Yun Guo}
\IEEEauthorblockA{\textit{Retail Platform Operation and} \\
\textit{Marketing Center, JD}\\
Beijing, China\\
yun96096@gmail.com}
\and
\IEEEauthorblockN{Wei Feng}
\IEEEauthorblockA{\textit{Retail Platform Operation and} \\
\textit{
Marketing Center, JD}\\
Beijing, China \\
fengwei25@jd.com}
\and
\IEEEauthorblockN{Zheng Zhang}
\IEEEauthorblockA{\textit{Retail Platform Operation and} \\
\textit{
Marketing Center, JD}\\
Beijing, China \\
zhangzheng11@jd.com}
\linebreakand
\IEEEauthorblockN{Xiancong Ren}
\IEEEauthorblockA{\textit{Retail Platform Operation and} \\
\textit{
Marketing Center, JD}\\
Beijing, China \\
renxiancong@jd.com}
\and
\IEEEauthorblockN{Yaoyu Li}
\IEEEauthorblockA{\textit{Retail Platform Operation and} \\
\textit{
Marketing Center, JD}\\
Beijing, China \\
liyaoyu1@jd.com}
\and
\IEEEauthorblockN{Jingjing Lv}
\IEEEauthorblockA{\textit{Retail Platform Operation and} \\
\textit{
Marketing Center, JD}\\
Beijing, China \\
lvjingjing1@jd.com}
\linebreakand
\IEEEauthorblockN{Xin Zhu}
\IEEEauthorblockA{\textit{Retail Platform Operation and} \\
\textit{
Marketing Center, JD}\\
Beijing, China \\
zhuxin3@jd.com}
\and
\IEEEauthorblockN{Zhangang Lin}
\IEEEauthorblockA{\textit{Retail Platform Operation and} \\
\textit{
Marketing Center, JD}\\
Beijing, China \\
linzhangang@jd.com}
\and
\IEEEauthorblockN{Jingping Shao}
\IEEEauthorblockA{\textit{Retail Platform Operation and} \\
\textit{
Marketing Center, JD}\\
Beijing, China \\
shaojingping@jd.com}
}

\maketitle

\begin{abstract}
Product image segmentation is vital in e-commerce. Most existing methods extract the product image foreground only based on the visual modality, making it difficult to distinguish irrelevant products. As product titles contain abundant appearance information and provide complementary cues for product image segmentation, we propose a mutual query network to segment products based on both visual and linguistic modalities. First, we design a language query vision module to obtain the response of language description in image areas, thus aligning the visual and linguistic representations across modalities. Then, a vision query language module utilizes the correlation between visual and linguistic modalities to filter the product title and effectively suppress the content irrelevant to the vision in the title. To promote the research in this field, we also construct a \textbf{M}ulti-\textbf{M}odal \textbf{P}roduct \textbf{S}egmentation dataset (MMPS), which contains 30,000 images and corresponding titles. The proposed method significantly outperforms the state-of-the-art methods on MMPS.
\end{abstract}


\section{Introduction}
\label{sec:intro}

\begin{figure}[t] 
\centering
\includegraphics[scale=0.20]{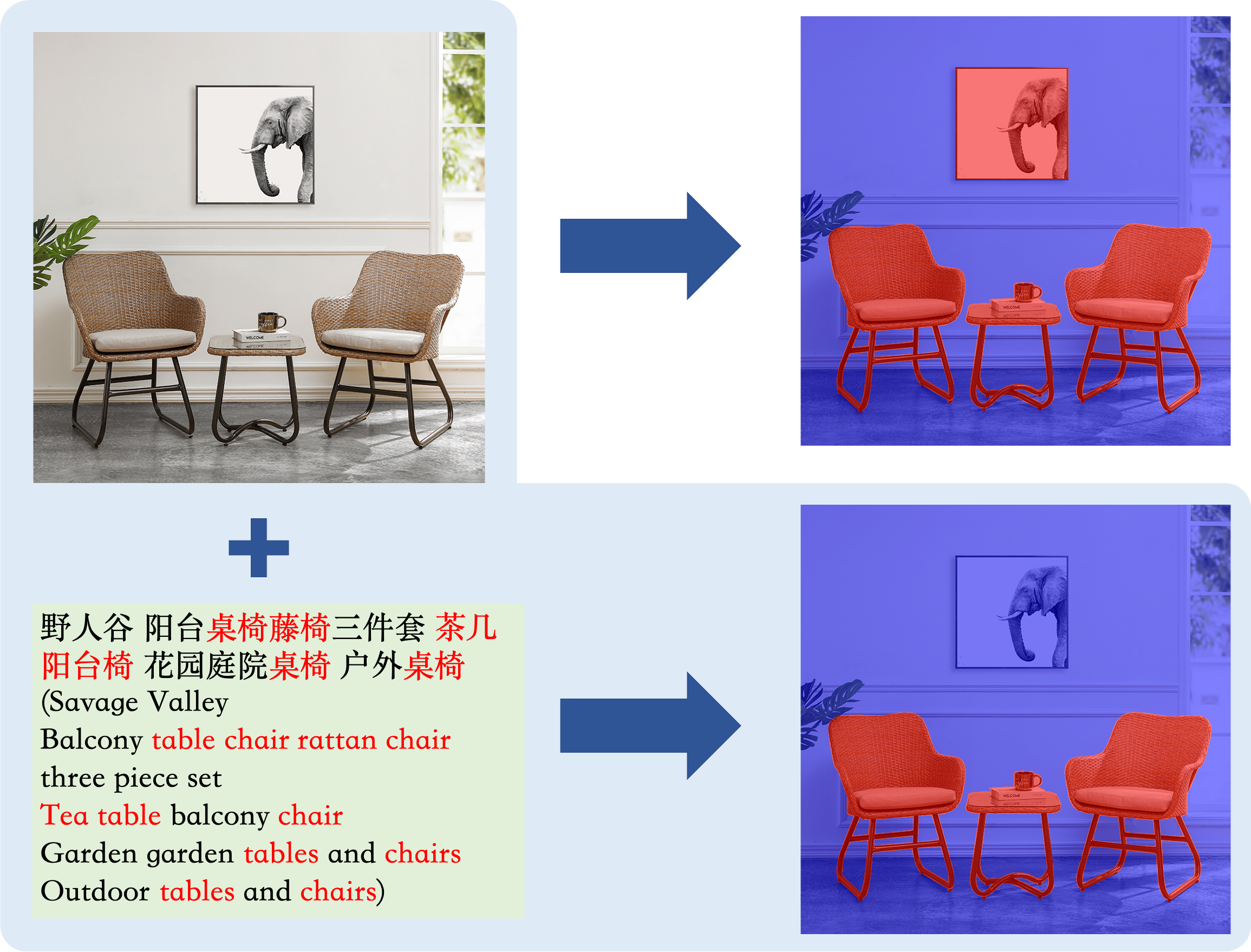}
\caption{
First row: Relying solely on the visual modality will lead to the misclassification of irrelevant objects into foreground. Second row: The introduction of product title information can help segment products more accurately. The content in black font in the title is visually irrelevant, and the English translation corresponding to the Chinese title is in brackets.
}
\label{fig:introduction}
\end{figure}

Product image segmentation aims to extract the foreground region of products in the image. It is widely used in advertising generation, image editing, and 2D virtual try-on. Despite the great progress made by previous works, it is still a challenging task due to the diversity of product appearance and the complexity of background.

Most previous methods \cite{joseph2019automatic,wu2021mr} adopt similar ideas to image segmentation and matting with some product-specific modifications. For example, Wu \emph{et al.} \cite{wu2021mr} deepened the network layers of the encoder and added residual blocks to improve the ability to extract details. Rajkumar \emph{et al.} \cite{joseph2019automatic} adopted a neural network classifier to identify if the background has a monocolor gradient. However, these methods only rely on the visual modality. When there are irrelevant objects around the target product, these irrelevant objects will be misclassified as foreground, as shown in the first row of Fig.~\ref{fig:introduction}.

In addition to the product image, the product title also contains abundant product appearance description information. Therefore, introducing linguistic modality could improve the quality of product image segmentation. As shown in the second row of Fig.~\ref{fig:introduction}, when the segmentation process refers to the linguistic modality, objects unrelated to the target product will be eliminated. Recently, some referring segmentation methods \cite{ding2021vision,shi2018key} have been proposed to segment the specified object in the image according to the text description. These methods provide inspiration for the fusion of product titles and images. However, compared with natural language description, the product title also contains noise irrelevant to the product appearance as shown in Fig.~\ref{fig:introduction}. Consequently, how to select the description related to appearance in the product title remains a problem.

In this paper, we propose a mutual query network, as shown in Fig.~\ref{fig:Module}, which can extract the appearance description information from the product title to enhance image segmentation. Specifically, we first adopt a language query vision module to obtain the response of the product title in the image area, which can generate a multi-modal feature map for each Chinese character in the product title. Then, based on the correlation of language and visual modalities, a vision query language module is proposed to filter content irrelevant to appearance in the multi-modal features. Finally, we densely integrate filtered multi-modal features into visual features to harvest accurate segmentation masks, as shown in Fig.~\ref{fig:framwork}. Besides, to promote the research in this field, we also construct a large \textbf{M}ulti-\textbf{M}odal \textbf{P}roduct \textbf{S}egmentation dataset (MMPS), which contains 30,000 images with corresponding titles. The links of MMPS dataset will be available at https://github.com/WeiFeng-Github/MQN.

In summary, the contributions of this paper are as follows:
\begin{itemize}
\item[$\bullet$] We propose a language query vision module, which uses the product title to help model focus on the target product in the image.
\item[$\bullet$] The proposed vision query language module uses the relevance of vision and language to filter the product title. The content irrelevant to the appearance in the title is effectively suppressed.
\item[$\bullet$] A large \textbf{M}ulti-\textbf{M}odal \textbf{P}roduct \textbf{S}egmentation dataset (MMPS) is constructed to promote the research of product image segmentation. We conduct extensive experiments on MMPS, and prove that the proposed method is superior to previous state-of-the-art semantic segmentation and referring segmentation methods.
\end{itemize} 


\section{Related Work}
\label{sec:rel}

\subsection{Product Image Segmentation}
As product image segmentation plays an important role in e-commerce, it has attracted increasing attention in recent years. Most previous methods segment products in a similar way to image segmentation and matting, where only the visual modality is used. To extract the apparel image mask, Zhu \emph{et al.} \cite{xinjuan2011apparel} adopted typical gradient formula to acquire the gradient matrix by computing the color gradient of the binary image. Rajkumar \emph{et al.} \cite{joseph2019automatic} designed a system that automatically removes background having monochrome gradients for retail product photography images. 
On the basis of UNet \cite{ronneberger2015u}, Wu \emph{et al.} \cite{wu2021mr} promoted the performance of product segmentation by deepening the network layers of the encoder and adding residual blocks. Li \emph{et al.} \cite{xie2021analysis} separated the products from the background through salient object detection.

Different from these methods, we utilize the complementary information from the product title, which enables the segmentation process to focus more on the target product. 

\subsection{Referring Image Segmentation}
Referring image segmentation \cite{ding2021vision,shi2018key} aims at generating fine segmentation masks from natural language descriptions. Since it also needs visual and linguistic modalities to generate masks, we review the relevant works here. Most existing methods extract visual and linguistic features respectively, then fuse multi-modal features to predict the segmentation mask. BRINet \cite{hu2020bi} used a bi-directional cross-modal attention module to learn the relationship between multi-modal features. Ding \emph{et al.} \cite{ding2021vision} produced multiple sets of queries, and adaptively selected the output features of these queries for better mask generation. LAVT \cite{yang2022lavt} proposed an early fusion scheme to integrate visual and linguistic features of multiple stages. Compared with expressions in referring image segmentation datasets \cite{yu2016modeling,mao2016generation}, the products are often in the prominent position of the image, so visual features are more critical. In addition, due to the noise in the product title, product segmentation needs to eliminate the noise in the product title as well as align the linguistic and visual modalities. However, the text description in referring image segmentation is cleaner, so they focus on the former.


\begin{figure*}[t] 
\centering
\includegraphics[scale=0.35]{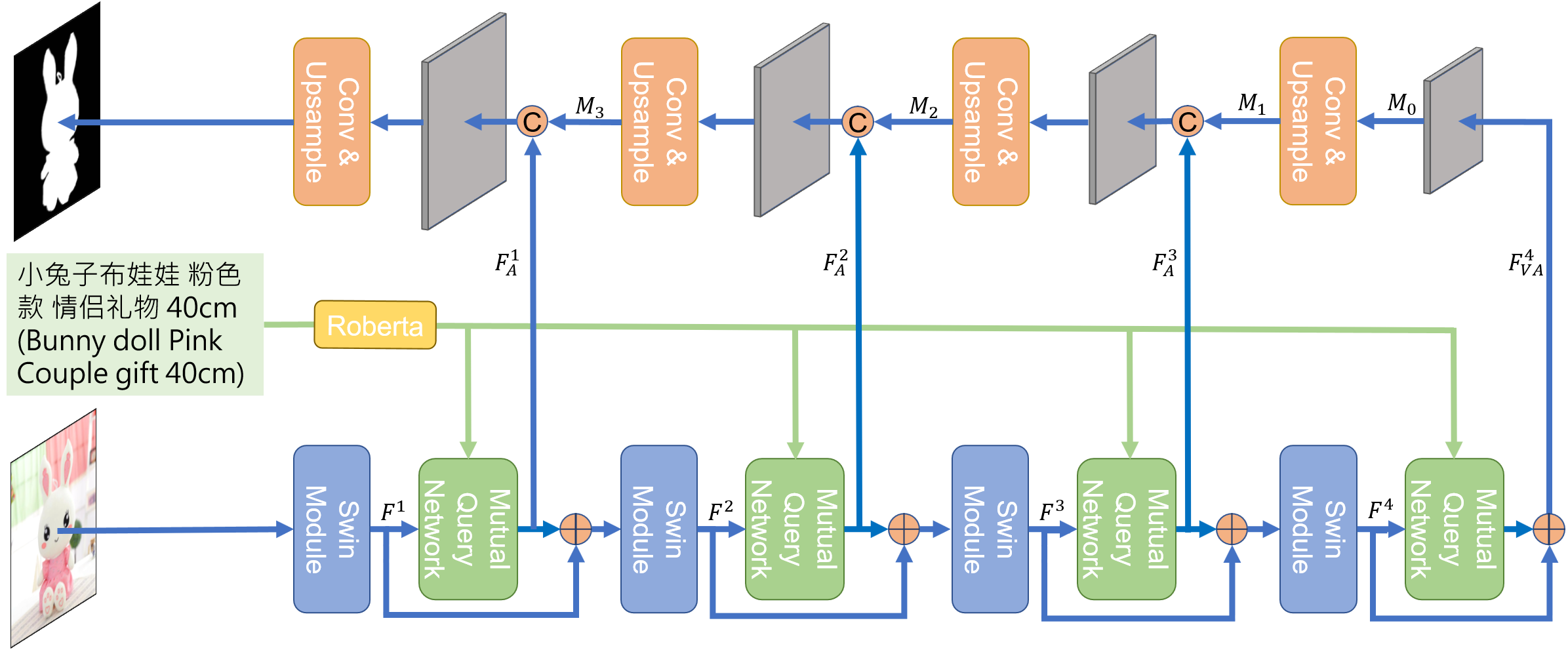}
\caption{
The overall framework of the proposed method. Here we show the product title in both Chinese and English.
}
\label{fig:framwork}
\end{figure*}

\section{Method}
\label{sec:Met}

The overall framework of our proposed method is shown in Fig.~\ref{fig:framwork}. Given a product image and the corresponding product description, we first use the pre-trained language model to extract the linguistic features from the product description and Swin Transformer \cite{liu2021swin} as a visual backbone to extract features at different stages. In each stage, we employ the proposed mutual query network to fuse extracted visual and linguistic features into multi-modal features, and filter out linguistic features irrelevant to vision. Four stacked stages achieve the dense integration of visual and linguistic features. In the end, the output features are converted into pixel-wise mask that delineates the product.

\subsection{Feature Extractors}

We briefly introduce the linguistic feature extractor Roberta \cite{liu2019roberta} and visual feature extractor Swin Transformer \cite{liu2021swin} and their training strategies.

\subsubsection{Visual Backbone} Swin Transformer has been widely used in various visual tasks \cite{cheng2022masked,cheng2021per} and achieved good results because of its powerful visual feature extraction ability. To extract visual features, the Swin Transformer first divides the input image into grids. Within its four swin modules, the network performs interaction between different pixel blocks through the Windows Multi-head Self-Attention (W-MSA) and the Shifted Windows Multi-head Self-Attention (SW-MSA), and finally generates four feature maps of different 
stages. We denote these features as ${F^i}$, where ${i=1,2,3,4}$.

\subsubsection{Linguistic Backbone} We use RoBERTa \cite{liu2019roberta} as our linguistic feature extractor, which improved the original BERT \cite{kenton2019bert} on the setting of many key hyperparameters and training strategies. To better extract the product title features in e-commerce scenarios, we collect 200 million product corpus from JD.com and design three domain-specific pretraining tasks: Masked Language Model (MLM), Attribute-Value Prediction (AVP) and Tertiary Category Prediction (TCP). Specifically, we first use Chinese Whole Word Mask (WWM) to tokenize the text, which masks a whole word for Chinese words, not for Chinese characters to make the model perceive the semantics of Chinese words. Then we train the linguistic feature extractor with the MLM objective. For the input product title, we mask some words randomly, and then input them into the language model to make the model predict the original sentence, which can be expressed as follows:

\begin{equation}
  {{L}_{MLM}}=-\sum\limits_{\hat{x}\in m\left( x \right)}{\log p\left( \hat{x}|{{X}_{/m\left( x \right)}} \right)},
\end{equation}
where ${X = \left\{ x_1, x_2, ..., x_T \right\}}$ means the product title; ${T}$ is the length of the sentence; ${m \left( x \right)}$ means the set of the masked words; ${{X}_{/m\left( x \right)}}$ demonstrates the rest words.

We also add AVP and TCP to predict the product value given the attribute and tertiary category that the product belongs to, to urge the model to learn more in-domain knowledge. AVP is dedicated to extracting product values from the product description based on product attribute queries, which can be formulated as:

\begin{equation}
  {{L}_{AVP}}=-\sum\limits_{a\in {{A}_{set}}}{\log p\left( a|X, A_Q \right)},
\end{equation}
where ${A_Q}$ is the attribute query; ${A_{set}}$ denotes the preset attribute value set for extracting from the description.

TCP is employed to judge the product category according to the product description, which is denoted as follows:

\begin{equation}
  {{L}_{TCP}}=-\sum\limits_{c\in {{C}_{set}}}{\log p\left( c|X \right)},
\end{equation}
where ${C_{set}}$ denotes the preset category set for product classification.

Notably, the above three pretraining tasks are combined for training. The training objective ${{L}_{Pre}}$ can be formulated as:
\begin{equation}
  {{L}_{Pre}}={{L}_{MLM}}+{{L}_{AVP}}+{{L}_{TCP}}.
\end{equation}

\subsection{Mutual Query Network}

\begin{figure}[t]
\centering
\includegraphics[height=40mm]{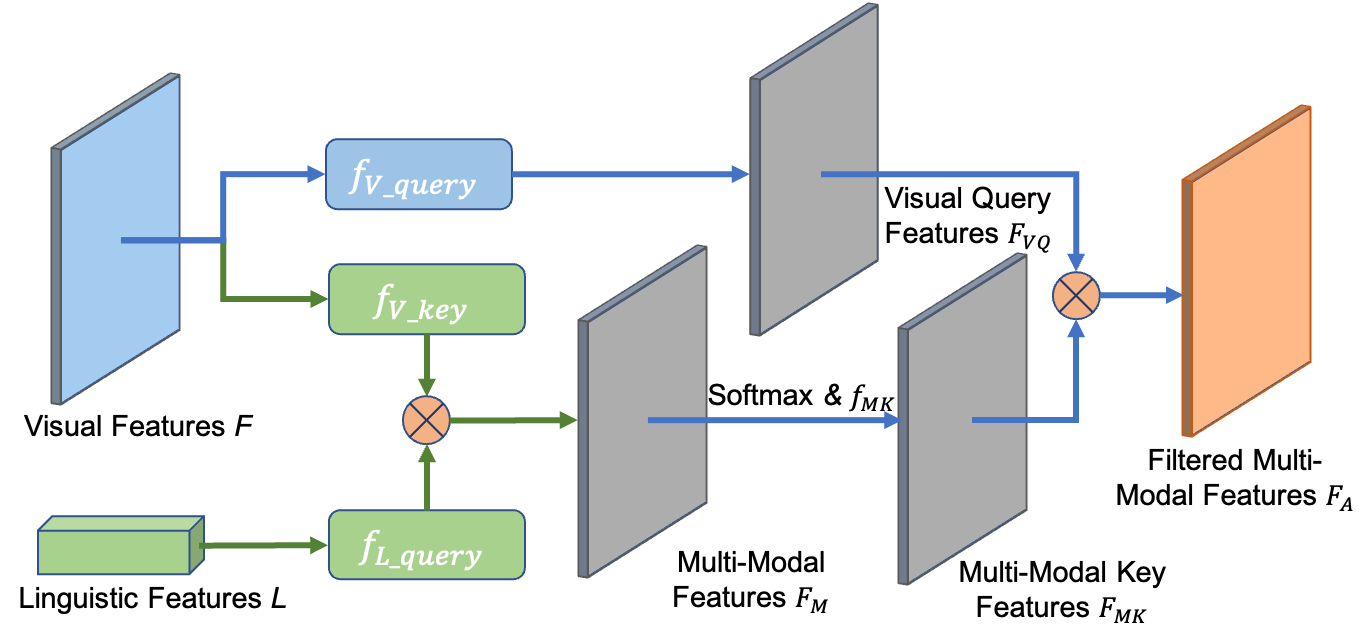}
\caption{
Pipeline of the proposed Mutual Query Network. The green lines and the blue lines represent the procedures of the language query vision module and the vision query language module respectively.
}
\label{fig:Module}
\end{figure}

The difference between the product description and the texts in referring image segmentation is in two folds. First, the product description contains more information about the product attributes represented by several key Chinese characters. Second, product description contains a large amount of redundant information visually agnostic, which will cause various interference to product segmentation. To 
incorporate useful information and avoid interference, we propose the vision-language mutual query network. The detailed structure is shown in Fig.~\ref{fig:Module}, which consists of two modules: Language Query Vision Module and Vision Query Language Module.

\subsubsection{Language Query Vision Module} In this step, we first convert linguistic features ${L \in R^{T \times C_{L}}}$, where ${T}$ is the length of the description and ${C_{L}}$ is the embedding dimension of words, into language query features ${{F_{L\_query}} \in R^{T \times C_{L}}}$ through language query projection ${f_{L\_query} : R^{C_L} \to R^{C_L}}$, and convert visual features ${F^i \in R^{H \times W \times C_{V}}}$ for a specific level ${i}$, where ${C_{V}}$ is the dimension of the visual features, into visual key features ${{F_{V\_key}} \in R^{H \times W \times C_{L}}}$ through visual key projection ${f_{V\_key} : R^{C_V} \to R^{C_L}}$. For simplicity, we omit the superscript ${i}$.

\begin{equation}
  {F_{L\_query}} = {{f_{L\_query}} \left( L \right) },
  {F_{V\_key}} = {{f_{V\_key}} \left( F \right) },
\end{equation}
where ${{f_{L\_query}}}$ and ${f_{V\_key}}$ are convolution operations.

Then language query features ${F_{L\_query}}$ and visual key features ${F_{V\_key}}$ are fused into multi-modal features ${F_{M} \in R^{H \times W \times T}}$.
\begin{equation}
  {F_{M}} = {F_{V\_key}} \cdot {{F_{L\_query}}^T},
\end{equation}
where ${\cdot}$ means dot product.

\subsubsection{Vision Query Language Module} In this step, we transform visual features ${F \in R^{H \times W \times C_{V}}}$ into visual query features ${F_{VQ} \in R^{H \times W \times C_{V}}}$ through visual query projection ${f_{V\_query} : R^{C_{V}} \to R^{C_{V}}}$.

\begin{equation}
  {F_{VQ}} = {f_{V\_query} \left( F \right) }.
\end{equation}

Then, multi-modal features ${F_{M}}$ are transformed into multi-modal key features through a softmax operation and a linear layer ${f_{MK} : R^{T} \to R^{C_V}}$.
\begin{equation}
   {F_{MK}} = f_{MK} \left( softmax \left( {F_M} \right) \right).
\end{equation}

We filter the multi-modal key features ${F_{MK}}$ through the visual query features ${F_{VQ}}$, and finally obtain the filtered multi-modal features ${F_A \in R^{H \times W \times C_{V}}}$.

\begin{equation}
  {F_A} = {F_{VQ}} \otimes F_{MK},
\end{equation}
where ${\otimes}$ stands for element-wise/hadamard product.

To fuse visual and textual information at different scales more fully, we adopt the mutual query network to fuse multiple visual features of different scales with linguistic features.

\subsection{Mask Prediction}

Based on the filtered multi-modal features ${F_A}$, we first fuse them into the input features of the swin modules in the backbone to acquire linguistic awareness. Then we design a series of convolution layers to fuse ${F_A}$ and the decoded features and upscale them to the largest scale. Finally, we decode the output feature into a product mask through a convolution layer.

Suppose the output feature of the ${i}$-th (${i \in \{1,2,3,4\}}$) swin module is ${F^{i}}$, and the output feature of the ${i}$-th mutual query network is ${F^{i}_A}$, the input feature of the ${i}$-th swin module is

\begin{equation}
  {F^{i}_{VA}} = {F^{i}} \oplus {F^{i}_A},
\label{eq:add}
\end{equation}
where ${\oplus}$ means element-wise addition.

In the ${j}$-th (${j \in \{1,2,3,4\}}$) decoding stage, we fuse the ${j-1}$-th decoded features $M_{j-1}$ and the output ${F^{5-j}_A}$ of ${5-j}$-th mutual query network by concatenation, upscale it, and perform convolution to generate output features, which can be formulated as follows.

\begin{equation}
  {M_{j}} = Conv \left( Upsample \left( \left[ {M_{j-1}}; {F^{5-j}_A} \right] \right) \right),
 \label{eq:cat}
\end{equation}
where ${Upsample()}$ means the upscale operation; ${Conv()}$ means convolution operation; ${\left[ ; \right]}$ stands for the concatenation operation; ${M_0} = {{F_{VA}^4}}$. In the last decoding layer, we decode it into a product mask ${M_4} \in R^{H \times W \times 2}$.

In training, we deploy the cross entropy loss ${CE \left( \right)}$ for a supervision.

\begin{equation}
  {Loss} = CE \left( \hat{M}, M_4 \right),
\end{equation}
where ${\hat{M}}$ is the ground truth mask for the image.

\subsection{Multi-Modal Product Segmentation Dataset}

\begin{figure}[h]
\centering
\includegraphics[height=51mm]{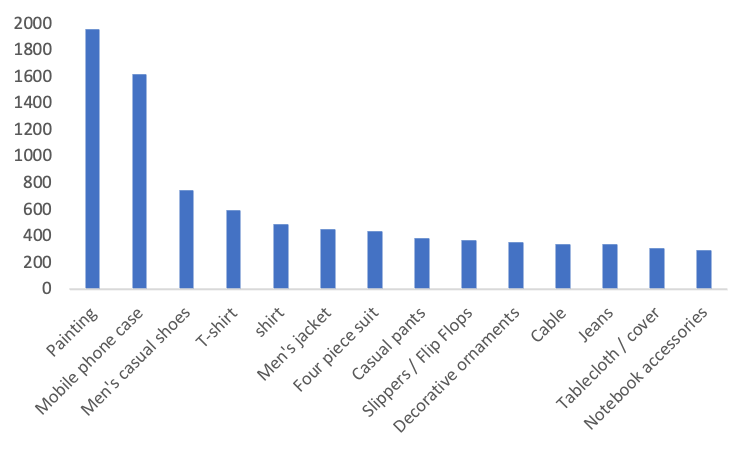}
\caption{
Product categories with more than 250 images in MMPS dataset.
}
\label{fig:mmps_label}
\end{figure}

\begin{figure}[t]
\centering
\includegraphics[height=65mm]{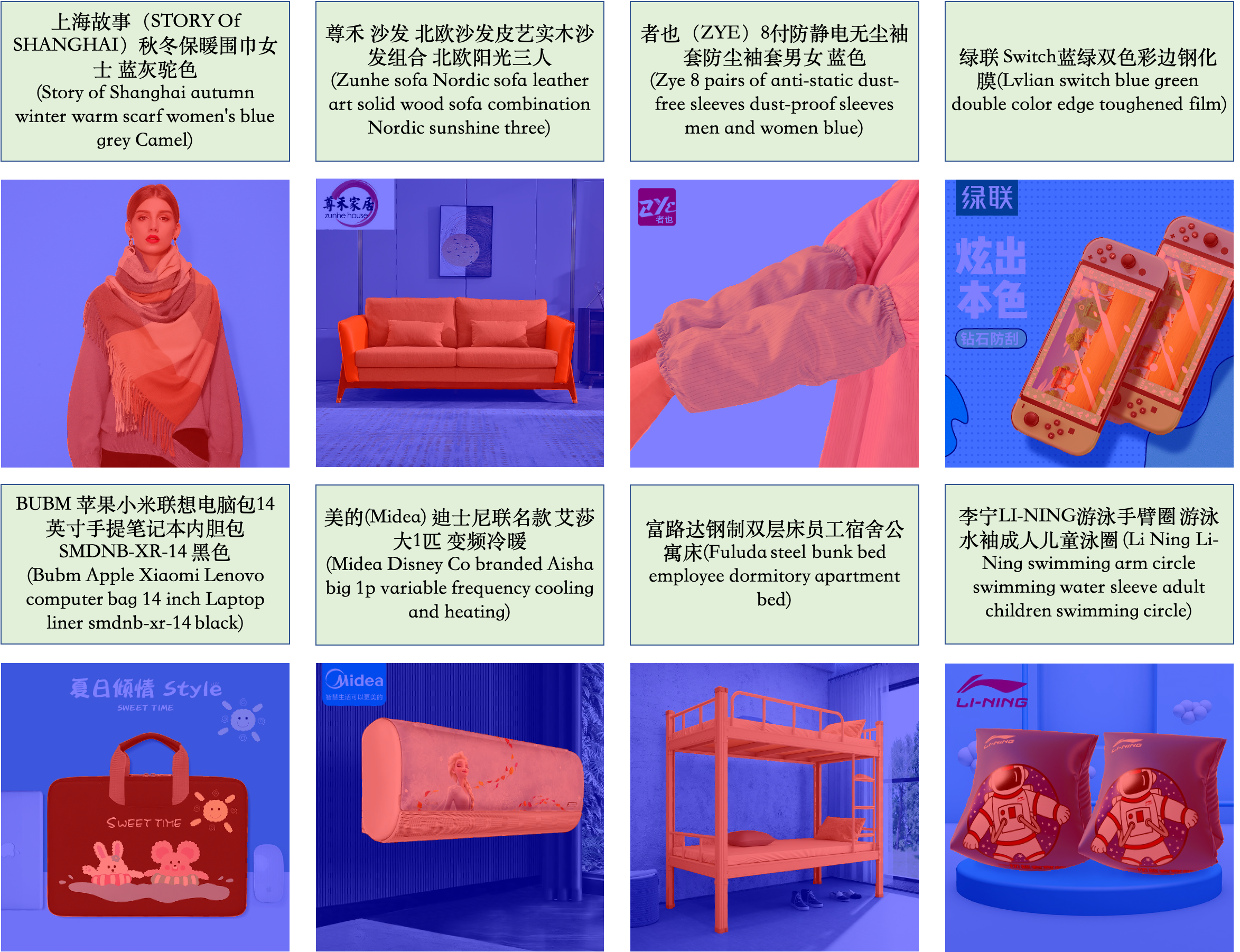}
\caption{
Examples of MMPS dataset. The red area is the annotation result. Here we show the product title in both Chinese and English.
}
\label{fig:mmps}
\end{figure}

The images of the Multi-Modal Product Segmentation dataset are collected from JD.com. The whole dataset contains 30,000 product images, of which 25,000 are used for training and 5,000 for testing. Most of them have a size of 800$\times$800. The products in this dataset cover 1,556 different categories, and we show the product categories with more than 250 images in Fig.~\ref{fig:mmps_label}. In order to satisfy segmentation, we annotate the foreground and background of the image. The corresponding product areas are labeled as the foreground, and the irrelevant areas are labeled as the background. We show some product images, corresponding titles, and annotations in Fig.~\ref{fig:mmps}.

We also extract the title of the product as its description, so as to provide text clues. The length of the product description varies from 0 to 100. For convenience, we will publish our pretrained language model. 

For evaluation, we adopt the same metrics as referring image segmentation to calculate the overall intersection-over-union (oIoU), mean intersection-over-union (mIoU), and precision at 0.5, 0.7, and 0.9 threshold values. For oIoU, we summarize all the intersection areas and union areas across all images, and calculate their quotients. For mIoU, we first calculate the IoU of each image, and then calculate the average value of all images.


\begin{figure*}[t]
\centering
\includegraphics[height=45mm]{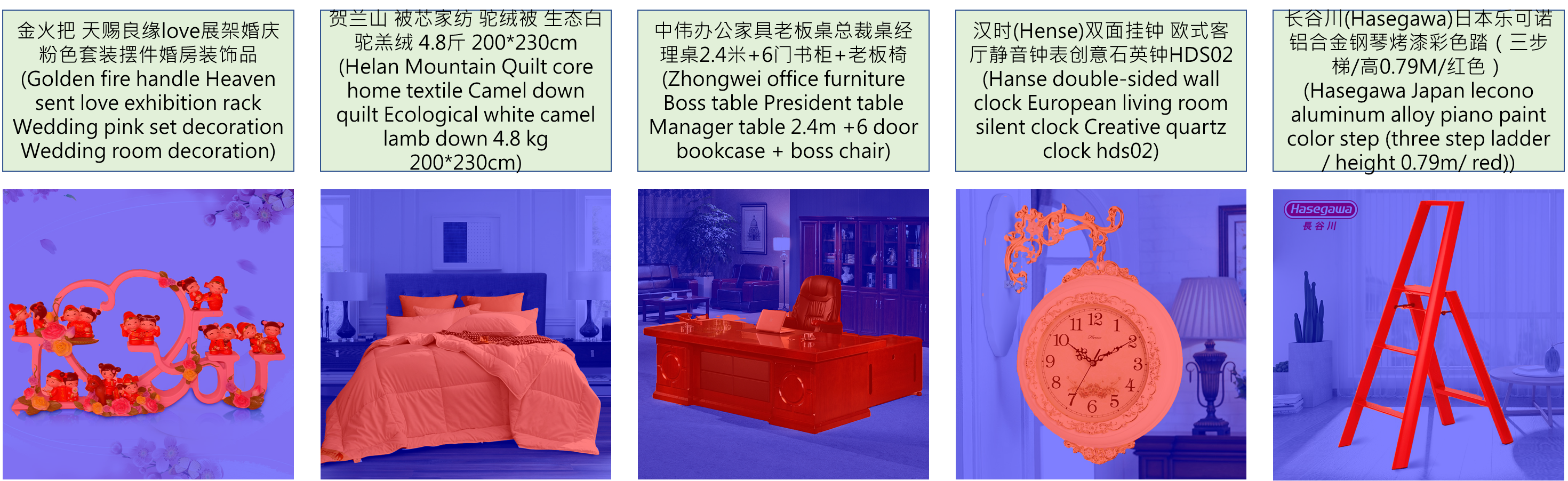}
\caption{
Qualitative examples of the proposed method. The red area is the output mask. Here we show the product title in both Chinese and English.
}
\label{fig:result}
\end{figure*}

\section{Experiments}
\label{sec:Exp}

\subsection{Implement Details} 

Our framework is implemented on PyTorch framework. The language model is pretrained in accordance with Sect 3.1. Swin Transformer base is adopted as the backbone of our vision model, which inherits parameters trained on ImageNet dataset \cite{deng2009imagenet}. The decoder and other weight parameters are initialized randomly. For the text descriptions, we first pad them to a fixed length of ${T=100}$, which are then embedded with a dim of ${C_L=300}$. During training, all input images are resized to 480$\times$480. We train the model by using the AdamW \cite{loshchilov2018decoupled} optimizer with a learning rate of 0.00005 for 30 epochs. All the experiments are conducted with a batch size of 8 on 4 GPUs. The implementation is on a workstation with a 2.40GHz 56-core CPU, 256G RAM, GTX Tesla P40, and 64-bit Red Hat.

\subsection{Comparison with State-of-the-art Methods}

In order to illustrate the superiority of our method, we implement several semantic and referring segmentation methods on the MMPS dataset, train and test them under the same conditions as our method.

\setlength{\tabcolsep}{4pt}
\begin{table}[t]
\begin{center}
\caption{
Results on MMPS test set. * indicates that the method uses visual and linguistic modalities.
}
\vspace{0.3cm}
\label{table:headings}
\begin{tabular}{llllll}
\hline\noalign{\smallskip}
Methods & P@0.5 & P@0.7 & P@0.9 & oIoU & mIoU\\
\hline\noalign{\smallskip}
FCN \cite{long2015fully}  & 93.82 & 84.50 & 48.62 & 83.02 & 82.97\\
Deeplab v3 \cite{chen2017rethinking} & 94.96 & 87.08 & 51.66 & 84.28 & 84.50\\
Pointrend \cite{kirillov2020pointrend} & 94.14 & 86.02 & 54.40 & 84.04 & 84.21\\
SETR \cite{zheng2021rethinking} & 94.78 & 89.02 & 56.66 & 85.55 & 85.08\\
Segmenter \cite{strudel2021segmenter} & 95.34 & 90.36 & 59.38 & 86.41 & 80.36\\
Segformer \cite{xie2021segformer} & 95.10 & 90.96 & 70.68 & 87.71 & 87.88\\
VLT* \cite{ding2021vision} & 94.78 & 89.32 & 62.36 & 86.32 & 85.76 \\
ReSTR* \cite{kim2022restr} & 94.88 & 90.12 & 68.66 & 87.88 & 88.12 \\
LAVT* \cite{yang2022lavt} & 95.86 & 92.46 & 74.88 & 88.84 & 89.18\\
\hline\noalign{\smallskip}
Ours* & {\bf{96.43}} & {\bf{93.82}} & {\bf{78.24}} & {\bf{89.26}} & {\bf{90.08}}\\
\hline
\end{tabular}
\end{center}
\end{table}
\setlength{\tabcolsep}{1.4pt}

\setlength{\tabcolsep}{4pt}
\begin{table}[t]
\begin{center}
\caption{
Results on RefCOCO dataset.
}
\vspace{0.3cm}
\label{table:refcoco}
\begin{tabular}{llll}
\hline\noalign{\smallskip}
Methods & val & test A  & test B\\
\hline\noalign{\smallskip}
CMSA \cite{ye2019cross} & 58.32 & 60.61 & 55.09 \\
BRINet \cite{hu2020bi} & 60.98 & 62.99 & 59.21  \\
VLT \cite{ding2021vision} & 65.65 & 68.29 & 62.73 \\
ReSTR \cite{kim2022restr} & 67.22 & 69.30 & 64.45 \\
LAVT \cite{yang2022lavt} & 72.73 & 75.82 & 68.79  \\
\hline\noalign{\smallskip}
Ours & {\bf{73.02}} & {\bf{76.29}} & {\bf{69.74}} \\
\hline
\end{tabular}
\end{center}
\end{table}
\setlength{\tabcolsep}{1.4pt}

As shown in Table.~\ref{table:headings}, we demonstrate the best performance of these methods and compare them with our method. Different from semantic segmentation methods, text clues provide our method with more product information, so that the model can better distinguish objects with the same visual significance. Benefiting from that, the mutual query network can reduce the impact of noise in titles, and our method also outperforms existing referring segmentation methods. We further evaluate the performance on a referring segmentation dataset RefCOCO \cite{yu2016modeling} as shown in Table.~\ref{table:refcoco}. Our method outperforms state-of-the-art referring segmentation approaches, which shows the superiority of our method. Some results are visualized as shown in Fig.~\ref{fig:result}. These results qualitatively illustrate the effectiveness and superiority of our method.

\subsection{Ablation Studies}

In this section, we conduct two ablation experiments to illustrate the effectiveness of language query vision and vision query language modules.

\subsubsection{Language Query Vision Module}
To demonstrate the benefits of linguistic features on product image segmentation, we evaluate a variant of our method that only relies on the visual modality. Without the help of product titles, the baseline model has difficulty dealing with complex scenes. As shown in Table.~\ref{table:ablation}, adding linguistic features, the model with the language query vision module outperforms the baseline model by 4.06\% (87.96\% vs 83.90\%) on oIoU and 4.6\% (88.48\% vs 83.88\%) on mIoU. Due to linguistic features providing effective clues for the segmentation process, the performance of product image segmentation is improved obviously. 


\setlength{\tabcolsep}{4pt}
\begin{table}[t]
\begin{center}
\caption{
Ablation studies on MMPS test set. w/ lqv is short for with language query vision module, and w/ vql is short for with vision query language module.
}
\vspace{0.3cm}
\label{table:ablation}
\begin{tabular}{lllllll}
\hline\noalign{\smallskip}
w/ lqv & w/ vql & P@0.5 & P@0.7 & P@0.9 & oIoU & mIoU\\
\noalign{\smallskip}
\hline
\noalign{\smallskip}
\XSolidBrush & \XSolidBrush & 93.40 & 85.88 & 56.20 & 83.90 & 83.88\\
\checkmark & \XSolidBrush & 96.04 & 91.60 & 71.20 & 87.96 & 88.48\\
\checkmark & \checkmark & {\bf{96.43}} & {\bf{93.82}} & {\bf{78.24}} & {\bf{89.26}} & {\bf{90.08}}\\
\hline
\end{tabular}
\end{center}
\end{table}
\setlength{\tabcolsep}{1.4pt}

\subsubsection{Vision Query Language Module} Although the product title contains rich descriptions of appearance, it also contains a lot of content unrelated to vision, which will damage the quality of segmentation. We show the effectiveness of the vision query language module by comparing with the model without it. The model without the vision query language module replaces ${F^{i-1}_A}$ in Eq.~\ref{eq:add} and Eq.~\ref{eq:cat} with the multi-modal features $F^{i-1}_M$. As shown in Table.~\ref{table:ablation}, the full model outperforms the model without the vision query language module 1.3\% (89.26\% vs 87.96\%) on oIoU and 1.6\% (90.08\% vs 88.48\%) on mIoU. These results prove that the content irrelevant to vision in the product title will introduce noise to the segmentation result, and the proposed vision query language module can effectively select linguistic features useful for segmentation.



\section{Conclusion}
\label{sec:Con}

In this paper, we propose a mutual query network to segment product images, in which product titles are introduced to help get accurate masks. Different from the previous methods only rely on the visual modality, we extract linguistic features beyond visual features and integrate them fully by language query vision module and vision query language module. Besides, to promote research in this field, we build a large multi-modal product segmentation dataset, named MMPS, containing abundant images with corresponding titles. We prove that the proposed method significantly outperforms the existing methods. We also conduct some ablation studies to demonstrate the effectiveness of language query vision and vision query language modules.

\bibliographystyle{IEEEtran}
\bibliography{IEEEabrv,icme2023}

\begin{thebibliography}{10}
\providecommand{\url}[1]{#1}
\csname url@samestyle\endcsname
\providecommand{\newblock}{\relax}
\providecommand{\bibinfo}[2]{#2}
\providecommand{\BIBentrySTDinterwordspacing}{\spaceskip=0pt\relax}
\providecommand{\BIBentryALTinterwordstretchfactor}{4}
\providecommand{\BIBentryALTinterwordspacing}{\spaceskip=\fontdimen2\font plus
\BIBentryALTinterwordstretchfactor\fontdimen3\font minus
  \fontdimen4\font\relax}
\providecommand{\BIBforeignlanguage}[2]{{%
\expandafter\ifx\csname l@#1\endcsname\relax
\typeout{** WARNING: IEEEtran.bst: No hyphenation pattern has been}%
\typeout{** loaded for the language `#1'. Using the pattern for}%
\typeout{** the default language instead.}%
\else
\language=\csname l@#1\endcsname
\fi
#2}}
\providecommand{\BIBdecl}{\relax}
\BIBdecl

\bibitem{joseph2019automatic}
R.~Joseph, N.~Naresh~Babu, R.~S. Murali \emph{et~al.}, ``Automatic retail
  product image enhancement and background removal,'' in \emph{AICC}, 2019, pp.
  1--15.

\bibitem{wu2021mr}
Z.~Wu, L.~Zhao, and H.~Zhang, ``Mr-unet commodity semantic segmentation based
  on transfer learning,'' \emph{IEEE Access}, vol.~9, pp. 159\,447--159\,456,
  2021.

\bibitem{ding2021vision}
H.~Ding, C.~Liu, S.~Wang \emph{et~al.}, ``Vision-language transformer and query
  generation for referring segmentation,'' in \emph{ICCV}, 2021, pp.
  16\,321--16\,330.

\bibitem{shi2018key}
H.~Shi, H.~Li, F.~Meng \emph{et~al.}, ``Key-word-aware network for referring
  expression image segmentation,'' in \emph{ECCV}, 2018, pp. 38--54.

\bibitem{xinjuan2011apparel}
Z.~Xinjuan, H.~Junfang, and Z.~Qianming, ``Apparel image matting and
  applications in e-commerce,'' in \emph{ITAIC}, vol.~2, 2011, pp. 278--282.

\bibitem{ronneberger2015u}
O.~Ronneberger, P.~Fischer, and T.~Brox, ``U-net: Convolutional networks for
  biomedical image segmentation,'' in \emph{MICCAI}, 2015, pp. 234--241.

\bibitem{xie2021analysis}
L.~Xie, ``Analysis of commodity image recognition based on deep learning,'' in
  \emph{ICMIP}, 2021, pp. 50--55.

\bibitem{hu2020bi}
Z.~Hu, G.~Feng, J.~Sun \emph{et~al.}, ``Bi-directional relationship inferring
  network for referring image segmentation,'' in \emph{CVPR}, 2020, pp.
  4424--4433.

\bibitem{yang2022lavt}
Z.~Yang, J.~Wang, Y.~Tang \emph{et~al.}, ``Lavt: Language-aware vision
  transformer for referring image segmentation,'' in \emph{CVPR}, 2022, pp.
  18\,155--18\,165.

\bibitem{yu2016modeling}
L.~Yu, P.~Poirson, S.~Yang \emph{et~al.}, ``Modeling context in referring
  expressions,'' in \emph{ECCV}, 2016, pp. 69--85.

\bibitem{mao2016generation}
J.~Mao, J.~Huang, A.~Toshev \emph{et~al.}, ``Generation and comprehension of
  unambiguous object descriptions,'' in \emph{CVPR}, 2016, pp. 11--20.

\bibitem{liu2021swin}
Z.~Liu, Y.~Lin, Y.~Cao \emph{et~al.}, ``Swin transformer: Hierarchical vision
  transformer using shifted windows,'' in \emph{ICCV}, 2021, pp.
  10\,012--10\,022.

\bibitem{liu2019roberta}
Y.~Liu, M.~Ott, N.~Goyal \emph{et~al.}, ``Roberta: A robustly optimized bert
  pretraining approach,'' \emph{arXiv preprint arXiv:1907.11692}, 2019.

\bibitem{cheng2022masked}
B.~Cheng, I.~Misra, A.~G. Schwing \emph{et~al.}, ``Masked-attention mask
  transformer for universal image segmentation,'' in \emph{CVPR}, 2022, pp.
  1290--1299.

\bibitem{cheng2021per}
B.~Cheng, A.~Schwing, and A.~Kirillov, ``Per-pixel classification is not all
  you need for semantic segmentation,'' \emph{NIPS}, vol.~34, pp.
  17\,864--17\,875, 2021.

\bibitem{kenton2019bert}
J.~D. M.-W.~C. Kenton and L.~K. Toutanova, ``Bert: Pre-training of deep
  bidirectional transformers for language understanding,'' in \emph{NAACL-HLT},
  2019, pp. 4171--4186.

\bibitem{deng2009imagenet}
J.~Deng, W.~Dong, R.~Socher \emph{et~al.}, ``Imagenet: A large-scale
  hierarchical image database,'' in \emph{CVPR}, 2009, pp. 248--255.

\bibitem{loshchilov2018decoupled}
I.~Loshchilov and F.~Hutter, ``Decoupled weight decay regularization,'' in
  \emph{ICLR}, 2018.

\bibitem{long2015fully}
J.~Long, E.~Shelhamer, and T.~Darrell, ``Fully convolutional networks for
  semantic segmentation,'' in \emph{CVPR}, 2015, pp. 3431--3440.

\bibitem{chen2017rethinking}
L.-C. Chen, G.~Papandreou, F.~Schroff \emph{et~al.}, ``Rethinking atrous
  convolution for semantic image segmentation,'' \emph{arXiv preprint
  arXiv:1706.05587}, 2017.

\bibitem{kirillov2020pointrend}
A.~Kirillov, Y.~Wu, K.~He \emph{et~al.}, ``Pointrend: Image segmentation as
  rendering,'' in \emph{CVPR}, 2020, pp. 9799--9808.

\bibitem{zheng2021rethinking}
S.~Zheng, J.~Lu, H.~Zhao \emph{et~al.}, ``Rethinking semantic segmentation from
  a sequence-to-sequence perspective with transformers,'' in \emph{CVPR}, 2021,
  pp. 6881--6890.

\bibitem{strudel2021segmenter}
R.~Strudel, R.~Garcia, I.~Laptev \emph{et~al.}, ``Segmenter: Transformer for
  semantic segmentation,'' in \emph{ICCV}, 2021, pp. 7262--7272.

\bibitem{xie2021segformer}
E.~Xie, W.~Wang, Z.~Yu \emph{et~al.}, ``Segformer: Simple and efficient design
  for semantic segmentation with transformers,'' \emph{NIPS}, vol.~34, pp.
  12\,077--12\,090, 2021.

\bibitem{kim2022restr}
N.~Kim, D.~Kim, C.~Lan \emph{et~al.}, ``Restr: Convolution-free referring image
  segmentation using transformers,'' in \emph{CVPR}, 2022, pp.
  18\,145--18\,154.

\bibitem{ye2019cross}
L.~Ye, M.~Rochan, Z.~Liu \emph{et~al.}, ``Cross-modal self-attention network
  for referring image segmentation,'' in \emph{CVPR}, 2019, pp.
  10\,502--10\,511.

\end{thebibliography}

\end{document}